\documentclass[10pt,twocolumn,letterpaper]{article}

\usepackage{cvpr}              

\usepackage{graphicx}
\usepackage{amsmath}
\usepackage{amssymb}
\usepackage{booktabs}
\usepackage{algorithm}  
\usepackage{algorithmicx}  
\usepackage{algpseudocode}  
\usepackage{multirow}
\usepackage{enumerate}
\usepackage{xcolor}
\usepackage{comment}

\usepackage[pagebackref,breaklinks,colorlinks]{hyperref}

\usepackage[capitalize]{cleveref}
\crefname{section}{Sec.}{Secs.}
\Crefname{section}{Section}{Sections}
\Crefname{table}{Table}{Tables}
\crefname{table}{Tab.}{Tabs.}

\usepackage{enumitem}
\setenumerate{itemsep=0pt,partopsep=0pt,parsep=0pt,topsep=2pt}

\def\confName{CVPR}
\def\confYear{2023}

\begin{document}

\title{Rethinking Out-of-distribution (OOD) Detection: \\Masked Image Modeling is All You Need}

\author{Jingyao Li$^{1}$ \quad\quad Pengguang Chen$^{2}$\quad\quad Zexin He$^{1}$\quad\quad Shaozuo Yu$^{1}$ \quad\quad Shu Liu$^{2}$ \quad\quad Jiaya Jia$^{1,2}$  \\[0.2cm]
 The Chinese University of Hong Kong$^{1}$\quad SmartMore$^{2}$\\
 jingyao.li@link.cuhk.edu.hk \quad  leojia@cse.cuhk.edu.hk
}

\maketitle

\begin{abstract}

The core of out-of-distribution (OOD) detection is to learn the in-distribution (ID) representation, which is distinguishable from OOD samples. Previous work applied recognition-based methods to learn the ID features, which tend to learn shortcuts instead of comprehensive representations. In this work, we find surprisingly that simply using reconstruction-based methods could boost the performance of OOD detection significantly. We deeply explore the main contributors of OOD detection and find that reconstruction-based pretext tasks have the potential to provide a generally applicable and efficacious prior, which benefits the model in learning intrinsic data distributions of the ID dataset. Specifically, we take Masked Image Modeling as a pretext task for our OOD detection framework (MOOD). Without bells and whistles, MOOD outperforms previous SOTA of one-class OOD detection by 5.7\%, multi-class OOD detection by 3.0\%, and near-distribution OOD detection by 2.1\%. It even defeats the 10-shot-per-class outlier exposure OOD detection, although we do not include any OOD samples for our detection. Codes are available at \href{}{https://github.com/JulietLJY/MOOD}.


\end{abstract}

\section{Introduction}
\label{sec:intro}

A reliable visual recognition system not only provides correct predictions on known context (also known as in-distribution data) but also detects unknown out-of-distribution (OOD) samples and rejects (or transfers) them to human intervention for safe handling. This motivates applications of outlier detectors before feeding input to the downstream networks, which is the main task of OOD detection, also referred to as novelty or anomaly detection. OOD detection is the task of identifying whether a test sample is drawn far from the in-distribution (ID) data or not. It is at the cornerstone of various safety-critical applications, including medical diagnosis \cite{caruana2015intelligible}, fraud detection \cite{phua2010comprehensive}, autonomous driving \cite{eykholt2018robust}, etc.

\begin{figure}[t]
  \centering
    \includegraphics[width=0.99\linewidth]{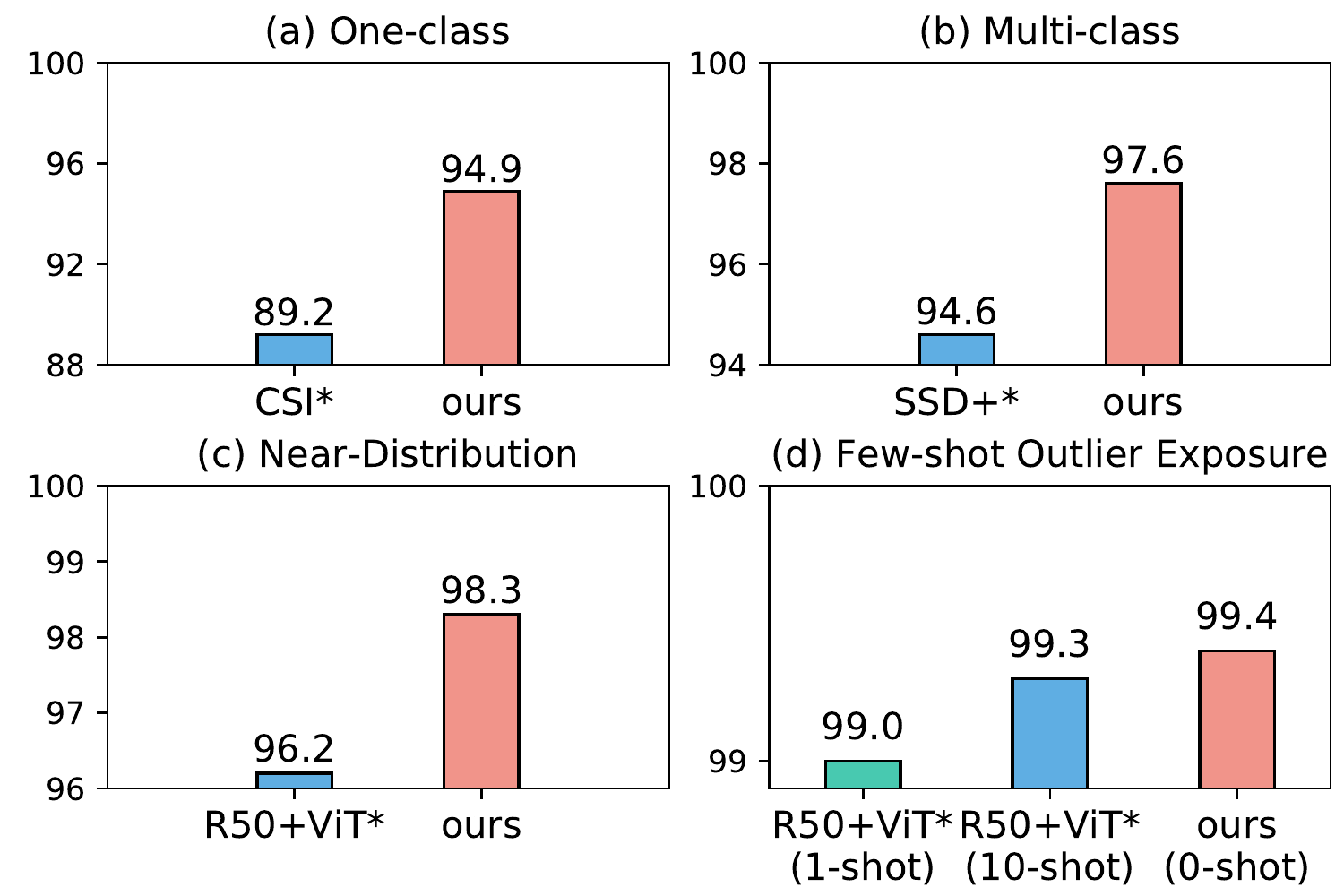}
    \caption{Performance of MOOD compared with current SOTA (indicated by `*') on four OOD detection tasks: (a) one-class OOD detection; (b) multi-class detection; (c) near-distribution detection; and (d) few-shot outlier exposure OOD detection.}
  \label{fig:performance}
\end{figure}

Many previous OOD detection approaches depend on outlier exposure \cite{ssd, oodlimits} to improve the performance of OOD detection, which turns OOD detection into a simple binary classification problem. We claim that the core of OOD detection is, instead, to learn the effective ID representation to discover OOD samples without any known outlier exposure. 

In this paper, we first present our surprising finding -- that is, {\it simply using reconstruction-based methods can {\it notably} boost the performance on various OOD detection tasks}. Our pioneer work along this line even outperforms previous few-shot outlier exposure OOD detection, albeit we do not include any OOD samples.

Existing methods perform contrastive learning \cite{csi, ssd} or pretrain classification on a large dataset \cite{oodlimits} to detect OOD samples. The former methods classify images according to the pseudo labels while the latter classifies images based on ground truth, whose core tasks are both to fulfill the classification target. However, research on backdoor attack \cite{backdoor_attack, frog_attack} shows that when learning is represented by classifying data, networks tend to take a shortcut to classify images. 

In a typical backdoor attack scene  \cite{frog_attack}, the attacker adds secret triggers on original training images with the visibly correct label. During the course of testing, the victim model classifies images with secret triggers into the wrong category. Research in this area demonstrates that networks only learn specific distinguishable patterns of different categories because it is a shortcut to fulfill the classification requirement. 

Nonetheless, learning these patterns is ineffective for OOD detection since the network does not understand the intrinsic data distribution of the ID images. Thus, learning representations by classifying ID data for OOD detection may not be satisfying. For example, when the patterns similar to some ID categories appear in OOD samples, the network could easily interpret these OOD samples as the ID data and classify them into the wrong ID categories. 

To remedy this issue, we introduce the reconstruction-based pretext task. Different from contrastive learning in existing OOD detection approaches \cite{csi, ssd}, our method forces the network to achieve the training purpose of reconstructing the image and thus makes it learn pixel-level data distribution. 

Specifically, we adopt the masked image modeling (MIM) \cite{bert, beit, mae} as our self-supervised pretext task, which has been demonstrated to have great potential in both natural language processing \cite{bert} and computer vision \cite{beit, mae}. In the MIM task, we split images into patches and randomly mask a proportion of image patches before feeding the corrupted input to the vision transformer. Then we use the tokens from discrete VAE \cite{tokenzier} as labels to supervise the network during training. With its procedure, the network learns information from remaining patches to speculate the masked patches and restore tokens of the original image. The reconstruction process enables the model to learn from the prior based on the intrinsic data distribution of images rather than just learning different patterns among categories in the classification process. 

In our extensive experiments, it is noteworthy that masked image modeling for OOD detection (MOOD) outperforms the current SOTA on all four tasks of one-class OOD detection, multi-class OOD detection, near-distribution OOD detection, and even few-shot outlier exposure OOD detection, as shown in \cref{fig:performance}. A few statistics are the following.

\begin{enumerate}
\item For one-class OOD detection (\cref{tab:one-class}), MOOD boosts the AUROC of current SOTA, i.e., CSI \cite{csi}, by \textbf{5.7\%} to \textbf{94.9\%}. 
\item For multi-class OOD detection (\cref{tab:multi-class}), MOOD outperforms current SOTA of SSD+ \cite{ssd} by \textbf{3.0\%} and reaches \textbf{97.6\%}. 
\item For near-distribution OOD detection (\cref{tab:structure}), AUROC of MOOD achieves \textbf{98.3\%}, which is \textbf{2.1\%} higher than the current SOTA of R50+ViT \cite{oodlimits}.
\item For few-shot outlier exposure OOD detection (\cref{tab:exposure}), MOOD (\textbf{99.41\%}) surprisingly defeats current SOTA of R50+ViT \cite{oodlimits} (with \textbf{99.29\%}), which makes use of 10 OOD samples per class. It is notable that we do not even include any OOD samples in MOOD.
\end{enumerate}

\section{Related Work}
\subsection{Out-of-distribution Detection}
A straightforward out-of-distribution (OOD) approach is to estimate the in-distribution (ID) density \cite{density_1,density_2,density_3,density_4} and reject test samples that deviate from the estimated distribution. Alternative methods base on the image reconstruction \cite{reconstruct_1, reconstruct_2, reconstruct_3}, learn the decision boundary between in- and out-of-distribution data \cite{boundary_1, boundary_2, boundary_3}, compute the distance between train and test features \cite{distance_1, distance_2, distance_3, csi, ssd}, etc..

In comparison, our work focuses on distance-based methods and yet includes the reconstruction-based methods as a pretext task. The key idea of distance-based approaches is that the OOD samples are supposedly far from the center of the in-distribution (ID) data \cite{ood_survey} in the feature space. Representative methods include K-nearest Neighbors \cite{distance_1}, prototype-based methods \cite{distance_2, distance_3}, etc.. We will explain the difference between our work and previous OOD detection methods later in this paper.

\subsection{Vision Transformer}
Transformer has achieved promising performance in computer vision \cite{beit, mae} and natural language processing \cite{bert}. Existing OOD detection research \cite{oodlimits} performs vision transformer (ViT  \cite{vit}) with classification pre-train on ImageNet-21k \cite{imagenet}. It mainly explores the impact of different structures on OOD detection tasks while we deeply explore the effect from four dimensions for OOD detection, including various pretext tasks, architectures, fine-tune processes, and OOD detection metrics. 

It is notable that extra OOD samples are utilized in various previous methods \cite{ssd, oodlimits} to further improve performance. In contrast, we argue that the exposure of OOD samples violates the original intention of OOD detection. In fact, a sufficient pretext task can achieve comparable or even superior results. Therefore, in our work, we focus on exploring an appropriate pretext task for OOD detection without including any OOD samples.

\begin{table*}[t]
\centering
\small
\setlength{\tabcolsep}{3.5mm}
\begin{tabular}{c|cccccccccccccc}
\toprule
In-Distribution          & \multicolumn{4}{c|}{CIFAR-10 $\longrightarrow $}                                    & \multicolumn{4}{c}{CIFAR-100 $\longrightarrow $}              \\
Out-of-Distribution         & SVHN          & CIFAR-100     & LSUN           & \multicolumn{1}{c|}{Avg}           & SVHN          & CIFAR-10      & LSUN          & Avg           \\
\midrule
Classification & 98.3          & 98.6          & 98.6           & \multicolumn{1}{c|}{98.5}          & 78.0          & 93.5          & 88.6          & 86.7          \\
MoCov3    & 98.6          & 92.4          & 89.8           & \multicolumn{1}{c|}{93.6}          & 78.8          & 72.8          & 75.8          & 75.8          \\
MIM         & \textbf{99.8} & \textbf{99.4} & \textbf{99.9}  & \multicolumn{1}{c|}{\textbf{99.7}} & \textbf{96.5} & \textbf{98.3} & \textbf{96.3} & \textbf{97.0} \\
\midrule
In-Distribution          & \multicolumn{8}{c}{ImageNet-30 $\longrightarrow $}                                                                                                  \\
Out-of-Distribution         & Dogs          & Places365     & Flowers102     & Pets                               & Food          & Dtd           & Caltech256    & Avg           \\
\midrule
Classification & \textbf{99.7} & 98.4          & 99.9           & \textbf{99.6}                      & \textbf{98.3} & 98.6          & 96.8          & 98.8          \\
MoCov3      & 88.2          & 82.0          & 99.3           & 81.1                               & 71.4          & 91.3          & 88.5          & 86.0          \\
MIM         & 99.4          & \textbf{98.9} & \textbf{100.0} & 99.1                               & 96.6          & \textbf{99.5} & \textbf{98.9} & \textbf{98.9} \\
\bottomrule
\end{tabular}
\caption{\textbf{Pretext Task}. AUROC (\%) of OOD detection on ViT with different pretext tasks on ImageNet22k. }
\label{tab:pretask}
\end{table*}

\subsection{Self-Supervised Pretext Task}
It has been long in the community to pre-train vision networks in various self-supervised manners, including generative learning \cite{pixelcnn, gpt, bert, beit}, contrastive learning \cite{moco, supcon, simclr, simsiam} and adversarial learning \cite{colorization, gan, adversial_ae}.
Among them, representative generative approaches include auto-regressive \cite{pixelcnn, gpt}, flow-based \cite{nice,glow}, auto-encoding \cite{bert, beit}, and hybrid generative methods \cite{graphaf, xlnet}. 

The self-supervised pretext task in our framework is Masked Image Modeling (MIM). It generally belongs to auto-encoding generative approaches. MIM was first proposed in natural language processing \cite{beit}. Its language modeling task randomly masks varying percentages of tokens of text and recovers the masked tokens from encoding results of the rest of text. Follow-up research \cite{bert, mae} transfers the similar idea from natural language processing to computer vision, masking different proportions of the image patches to recover results.

Multiple existing methods take advantage of self-supervised tasks to guide learning of representation for OOD detection. The latest work \cite{csi, ssd} presents contrastive learning models as feature extractors. However, existing approaches of classifying transformed images according to contrastive learning possess similar limitations -- that is, the model tends to learn the specific patterns of categories, which are beneficial for classification but do not help understand intrinsic data  distributions of ID images. 

Research of \cite{oodlimits} also mentioned this problem. However, the introduced large-scale pre-trained transformers \cite{oodlimits} may not jump out of the loop, in our observation, because the pretext task remained to be classification. In our work, we address this issue by performing the masked image modeling task for OOD detection.


\section{Method}
In this section, we first explain the main factors to help OOD detection and finally propose our framework to achieve this goal. 

We first define the notations. For a given dataset $X_{\rm ID}$, the goal of out-of-distribution (OOD) detection is to model a detector that identifies whether an input image $x \in X_{\rm ID}$ or $x \notin X_{\rm ID}$ (that is, $x \in X_{\rm OOD}$). A majority of existing methods for OOD detection define an OOD score function $s(x)$. Its abnormal high or low value represents that $x$ is from out-of-distribution. 

\subsection{Choosing the Pretext Task} 
\label{sec:pretask}

In this section, we choose the pretext task that can provide the intrinsic prior to suit the OOD detection task. Most previous OOD methods learn the ID representation through classification \cite{baseline_ood, oodlimits} or contrastive learning \cite{csi, ssd} on ID samples, which take advantage of either the ground truth or pseudo labels to supervise the classification networks. 

On the other hand, work of \cite{backdoor_attack, frog_attack} shows that classification networks only learn different patterns among training categories because it is a shortcut to fulfill classification. It is indicated that the network actually does not understand the intrinsic data distribution of the ID images.

In comparison, the reconstruction-based pretext task forces the network to learn the real data distribution of the ID images during training to reconstruct the image instead of the patterns for classification. Benefiting from these priors, the network can learn a more representative feature of the ID dataset. It enlarges the divergence between the OOD and ID samples. 

In our method, we pre-train the model with Masked Image Modeling (MIM) pretext \cite{bert} on a large dataset and fine-tune it on the ID dataset. We compare the performance of MIM and contrastive learning pretext task MoCov3 \cite{mocov3} in ~\cref{tab:pretask}. It shows that the performance of MIM is much increased by 13.3\% to 98.66\%. 

\subsection{Exploring Architecture} 
\label{sec:arch}
\begin{table}[t]
\small
\centering
\setlength{\tabcolsep}{3.5mm}
\begin{tabular}{c|ccccccccccc}
\toprule
\multirow{2}{*}{Model} & Fine-tuned          & \multirow{2}{*}{AUROC(\%)} \\
                       & Test Acc(\%)        &                            \\
\midrule
BiT R50 \cite{oodlimits} & 87.01               & 81.71                      \\
BiT R101$\times$3 \cite{oodlimits}& 91.55               & 90.10                      \\
ViT \cite{oodlimits} & 90.95               & 95.53                      \\
MLP-Mixer \cite{oodlimits} & 90.40               & 95.31                      \\
R50 + ViT (SOTA) \cite{oodlimits} & \textbf{91.71}               & \textbf{96.23}  \\
\bottomrule
\end{tabular}
\caption{\textbf{Architecture}. AUROC (\%) of OOD detection with various architectures. The last line shows our improvement. The ID and OOD datasets are CIFAR-100 and CIFAR-10, respectively. }
\label{tab:structure}
\end{table}

To explore an effective architecture \cite{oodlimits}, we evaluate OOD detection performance on BiT (Big Transfer \cite{bit}) and MLP-Mixer, in comparison with ViT. We adopt CIFAR-100 and CIFAR-10 \cite{cifar} as the ID-OOD pair. They have close distributions because of their similar semantics and construction. Results are in \cref{tab:structure}.

R50 + ViT \cite{vit, resnet} is the current SOTA on near-distribution OOD detection \cite{oodlimits}, which doubles the model size and testing time but achieves only 96.23\% (0.70\% higher than ViT). However, MIM on a single ViT significantly improves its AUROC to 98.30\% (2.07\% higher), without any additional source assumption. It manifests that efficient pretext itself is sufficient for producing distinguishable representation -- {\it there is no need to use a larger model or combination of multiple models} in this regard.

\subsection{About Fine-Tuning} 
\label{sec:fine-tune}

\begin{table}[t]
\small
\centering
\setlength{\tabcolsep}{2.1mm}
\begin{tabular}{c|ccc|ccccccc}
\toprule
One-Class                    & \multicolumn{3}{c|}{fine-tune}         & \multirow{2}{*}{AUROC(\%)} \\
Dataset                      & MIM-pt     & inter-ft   & fine-tune         & \\
\midrule
\multirow{2}{*}{CIFAR-10}    & \checkmark &            &            & 72.2 \\
                             & \checkmark & \checkmark &            & \textbf{97.9} \\
\midrule
\multirow{2}{*}{CIFAR-100}   & \checkmark &            &            & 66.3  \\
                             & \checkmark & \checkmark &            & \textbf{96.5} \\
\midrule
\multirow{2}{*}{ImageNet-30} & \checkmark &            &            & 75.2   \\
                             & \checkmark &            & \checkmark & \textbf{92.0} \\
\bottomrule
\end{tabular}
\caption{\textbf{Fine-tuning} (One-class). AUROC (\%) of OOD detection with different fine-tuning processes on one-class CIFAR-10, CIFAR-100 (super-classes) and ImageNet-30.}
\label{tab:1class-fine-tune}
\end{table}

\begin{table*}[t]
\centering
\small
\setlength{\tabcolsep}{3.2mm}
\begin{tabular}{ccc|ccccccccccccc}
\toprule
\multicolumn{3}{c}{finetune}                   & \multicolumn{4}{|c}{CIFAR-10 $\longrightarrow $}                             & \multicolumn{4}{|c}{CIFAR-100 $\longrightarrow $}                                   \\
MIM-pt               & inter ft   & ft         & SVHN           & CIFAR-100     & LSUN           & Avg                       & \multicolumn{1}{|c}{SVHN}          & CIFAR-10             & LSUN                 & Avg                  \\
\midrule
\checkmark           &            &            & 62.2           & 62.9          & 98.5           & \multicolumn{1}{c|}{74.5} & 48.4          & 42.2                 & 96.0                 & 62.2                 \\
\checkmark           & \checkmark &            & 89.5           & 90.0          & 99.8           & \multicolumn{1}{c|}{93.1} & 74.3          & 62.0                 & \textbf{98.3}        & 68.2                 \\
\checkmark           &            & \checkmark & 99.1           & 94.6          & 97.4           & \multicolumn{1}{c|}{97.0}   & 93.7          & 83.7                 & 91.4                 & 89.6                 \\
\checkmark           & \checkmark & \checkmark & \textbf{99.8}  & \textbf{99.4} & \textbf{99.9}  & \multicolumn{1}{c|}{\textbf{99.7}} & \textbf{96.5} & \textbf{98.3}        & 96.3                 & \textbf{97.0}        \\
\midrule
\multicolumn{3}{c}{finetune}                   & \multicolumn{8}{|c}{ImageNet30 $\longrightarrow $}                                                                                                                \\
MIM-pt               & inter-ft   & ft         & Dogs           & Places365     & Flowers102     & Pets                      & Food          & Caltech256           & Dtd                  & Avg                  \\
\midrule
\checkmark           &            &            & 60.2           & 82.7          & 28.6           & 41.9                      & 72.5          & 42.2                 & 29.4                 & 51.1                 \\
\checkmark           & \checkmark &            & \textbf{100.0} & 97.9          & 99.9           & \textbf{99.6}             & \textbf{97.1} & 96.9                 & 98.2                 & 98.2                 \\
\checkmark           &            & \checkmark & 91.3           & 97.0          & 95.1           & 93.8                      & 99.3          & 84.0                 & 95.4                 & 92.9                 \\
\checkmark           & \checkmark & \checkmark & 99.4           & \textbf{98.9} & \textbf{100.0} & 99.1                      & 96.6          & \textbf{99.5}        & \textbf{98.9}        & \textbf{98.9}       \\
\bottomrule
\end{tabular}
\caption{\textbf{Fine-tuning} (Multi-class). AUROC (\%) of OOD detection with different fine-tuning processes on multi-class CIFAR-10, CIFAR-100 and ImageNet-30. }
\label{tab:metric}
\label{tab:fine-tune}
\end{table*}

\begin{table*}[t]
\centering
\small
\setlength{\tabcolsep}{3.5mm}
\begin{tabular}{c|ccccccccccccccc}
\toprule
In-Distribution                           & \multicolumn{4}{c|}{CIFAR-10 $\longrightarrow $}                                                        & \multicolumn{4}{c}{CIFAR-100 $\longrightarrow $}                                          \\
Out-of-Distribution                          & SVHN                 & CIFAR-100            & LSUN                 & \multicolumn{1}{c|}{Avg}           & SVHN                 & CIFAR-10             & LSUN                 & Avg                  \\
\midrule
Softmax  \cite{baseline_ood} & 88.6                 & 85.8                 & 90.7                 & \multicolumn{1}{c|}{88.4}          & 81.9                 & 81.1                 & 86.6                 & 83.2                 \\
Entropy  \cite{baseline_ood} & \textbf{99.9}        & 97.1                 & 98.1                 & \multicolumn{1}{c|}{98.4}          & 93.7                 & 94.1                 & 88.7                 & 92.2                 \\
Energy   \cite{energy}       & \textbf{99.9}        & 97.0                 & 97.6                 & \multicolumn{1}{c|}{98.2}          & 92.8                 & 93.5                 & 86.1                 & 90.8                 \\
GradNorm  \cite{gradnorm}    & 99.6                 & 94.3                 & 87.8                 & \multicolumn{1}{c|}{93.9}          & 61.6                 & 87.7                 & 38.4                 & 62.6                 \\
Distance  \cite{mahalanobis} & 99.8                 & \textbf{99.4}        & \textbf{99.9}        & \multicolumn{1}{c|}{\textbf{99.7}} & \textbf{96.5}        & \textbf{98.3}        & \textbf{96.3}        & \textbf{97.0}        \\
\midrule
In-Distribution                           & \multicolumn{8}{c}{ImageNet-30 $\longrightarrow $}                                                                                                                                                  \\
Out-of-Distribution                          & Dogs                 & Places365            & Flowers102           & Pets                               & Food                 & Dtd                  & Caltech256           & Avg                  \\
\midrule
Softmax  \cite{baseline_ood} & 96.7                 & 90.5                 & 89.7                 & 95.0                               & 79.8                 & 90.6                 & 90.1                 & 90.3                 \\
Entropy  \cite{baseline_ood} & 92.5                 & 87.2                 & 97.5                 & 90.6                               & 69.6                 & 94.9                 & 85.7                 & 88.3                 \\
Energy   \cite{energy}       & 89.7                 & 82.1                 & 95.8                 & 88.1                               & 67.8                 & 93.1                 & 82.3                 & 85.6                 \\
GradNorm  \cite{gradnorm}    & 74.8                 & 78.7                 & 92.0                 & 70.6                               & 61.5                 & 90.3                 & 74.3                 & 77.5                 \\
Distance  \cite{mahalanobis} & \textbf{99.4}        & \textbf{98.9}        & \textbf{100.0}       & \textbf{99.1}                      & \textbf{96.6}        & \textbf{99.5}        & \textbf{98.9}        & \textbf{98.9}       \\
\bottomrule
\end{tabular}
\caption{\textbf{Metric}. AUROC (\%) of OOD detection with different metrics on multi-class CIFAR-10, CIFAR-100 and ImageNet-30. }
\label{tab:metric}
\label{tab:fine-tune}
\end{table*}

\noindent\textbf{One-class Fine-tuning.} For one-class OOD detection, we pre-train the MIM model and finely tune it on ImageNet-21k \cite{imagenet}, as recommended by BEiT \cite{beit}. In particular, when performing one-class OOD detection on ImageNet-30, since we do not include the OOD labels during training, we only pre-train it on ImageNet-21k without intermediate fine-tuning. Therefore, we utilize the label smoothing \cite{ls} to help the model learn from the one-class fine-tune task on the ID dataset as
\begin{equation}\label{equ:ls}
    y_c^{LS} = y_c(1-\alpha)+\alpha/N_c, \quad\quad c=1, 2, \dots, N_c
\end{equation}
where $c$ is the index of category; $N_c$ is the number of classes; and $\alpha$ is the hyperparameter that determines smoothing level. If $\alpha=0$, we obtain the original one-hot encoded $y_c$ and if $\alpha=1$, we get the uniform distribution.

Label smoothing was used to address overfitting and overconfidence in normal fine-tuning process. We, instead, find that it can be utilized in one-class fine-tuning. The performance of the model before and after one-class fine-tune is illustrated in \cref{tab:1class-fine-tune}. It is clear that the model actually learns information from the one-class fine-tuning operation. This may be counter-intuitive because the labels are equal. The reason is, due to label smoothing, the loss is larger than 0 and persuades the model to update parameters, although the accuracy reaches 1.

\vspace{2mm}\noindent\textbf{Multi-class Fine-tuning.}
For multi-class OOD detection, we pre-train the MIM model, intermediately use fine-tuning on ImageNet-21k \cite{imagenet}, and apply fine-tuning again on the ID dataset. We perform experiments to validate the effectiveness of each stage in \cref{tab:fine-tune}. It proves that all stages contribute well to the performance of OOD detection.


\subsection{OOD Detection Metric is Important}
\label{sec:metric}

Here, we compare the performance of several commonly-used OOD detection metrics, including Softmax \cite{baseline_ood}, Entropy \cite{baseline_ood}, Energy \cite{energy}, GradNorm  \cite{gradnorm} and Mahalanobis distance \cite{mahalanobis}. We perform OOD detection with MIM pretext task with each metric -- the results are shown in \cref{tab:metric}. They prove that the Mahalanobis distance is a better metric for MOOD.

\subsection{Final Algorithm of MOOD} 
\label{sec:alg}

To sum up, in this section, we have explored the effect of contributors to OOD detection, including various pretext tasks, architectures, fine-tuning processes, and OOD detection metrics. In general, we find that the finely tuned MOOD on ViT with Mahalanobis distances achieves the best result. The outstanding performance of MOOD demonstrates that an efficient pretext task itself is sufficient for producing distinguishable representation, and there is no need for a larger model or multi-models. 

In \cref{sec:exp}, we will show that few-shot outlier exposure utilized in multiple existing OOD detection approaches \cite{ssd, oodlimits} is also unnecessary. The algorithm of MOOD is shown in the Appendix. It mainly includes the following stages.

\begin{enumerate}
\item Pre-train the Masked Image Modeling ViT on ImageNet-21k.
\item Apply intermediate fine-tuning ViT on ImageNet-21k. 
\item Apply fine-tuning of pre-trained ViT on the ID dataset. 
\item Extract features from the trained ViT and calculate the Mahalanobis distance metric for OOD detection.
\end{enumerate}

\begin{table*}[t]
\small
\centering
\begin{subtable}{0.99\linewidth}
\centering
\setlength{\tabcolsep}{1.5mm}
\begin{tabular}{c|ccccccccccccccc}
\toprule
Method   & Plane & Car   & Bird & Cat & Dear& Dog & Frog& Horse & Ship& Truck & Average  \\
\midrule
OC-SVM\cite{goad} & 65.6 & 40.9 & 65.3 & 50.1 & 75.2 & 51.2 & 71.8 & 51.2 & 67.9 & 48.5 & 58.8   \\
DeepSVDD\cite{deepsvdd}  & 61.7 & 65.9 & 50.8 & 59.1 & 60.9 & 65.7 & 67.7 & 67.3 & 75.9 & 73.1 & 64.8   \\
AnoGAN\cite{anogan} & 67.1 & 54.7 & 52.9 & 54.5 & 65.1 & 60.3 & 58.5 & 62.5 & 75.8 & 66.5 & 61.8   \\
OCGANOCGAN\cite{ocgan} & 75.7 & 53.1 & 64.0 & 62.0 & 72.3 & 62.0 & 72.3 & 57.5 & 82.0 & 55.4 & 65.7   \\
Geom\cite{geom}  & 74.7 & 95.7 & 78.1 & 72.4 & 87.8 & 87.8 & 83.4 & 95.5 & 93.3 & 91.3 & 86.0   \\
Rot\cite{rot}   & 71.9 & 94.5 & 78.4 & 70.0 & 77.2 & 86.6 & 81.6 & 93.7 & 90.7 & 88.8 & 83.3   \\
Rot+Trans\cite{rot} & 77.5 & 96.9 & 87.3 & 80.9 & 92.7 & 90.2 & 90.9 & 96.5 & 95.2 & 93.3 & 90.1   \\
GOAD\cite{goad}  & 77.2 & 96.7 & 83.3 & 77.7 & 87.8 & 87.8 & 90.0 & 96.1 & 93.8 & 92.0 & 88.2   \\
CSI (SOTA)\cite{csi}   & 89.9 & 99.1 & 93.1 & 86.4 & 93.9 & 93.2 & 95.1 & 98.7 & 97.9 & 95.5 & 94.3   \\
ours  & \textbf{98.6}\tiny{$\pm$0.4} & \textbf{99.3}\tiny{$\pm$0.5} & \textbf{94.3}\tiny{$\pm$0.6} & \textbf{93.2}\tiny{$\pm$0.5} & \textbf{98.1}\tiny{$\pm$0.6} & \textbf{96.5}\tiny{$\pm$0.4} & \textbf{99.3}\tiny{$\pm$0.2} & \textbf{99.0}\tiny{$\pm$0.1} & \textbf{98.8}\tiny{$\pm$0.1} & \textbf{97.8}\tiny{$\pm$0.4} & \textbf{97.8}\tiny{$\pm$0.4} \\
(improve)   & {\color{teal}+8.7}& {\color{teal}+0.2}& {\color{teal}+1.2}& {\color{teal}+6.8}& {\color{teal}+4.2}& {\color{teal}+3.3}& {\color{teal}+4.2}& {\color{teal}+0.3}& {\color{teal}+0.9}& {\color{teal}+2.3}& {\color{teal}+3.5} \\
\bottomrule
\end{tabular}
\caption{CIFAR-10}
\label{tab:one-class-cifar10}
\end{subtable}
\\
\begin{subtable}{0.48\linewidth}
\centering
\setlength{\tabcolsep}{12mm}
\begin{tabular}{c|ccccccccccc}
\toprule
Method & AUROC   \\
\midrule
OC-SVM\cite{goad} & 63.1  \\
Geom\cite{geom}  & 78.7  \\
Rot\cite{rot}   & 77.7  \\
Rot+Trans\cite{rot} & 79.8  \\
GOAD\cite{goad}  & 74.5  \\
CSI (SOTA)\cite{csi}    & 89.6  \\
ours  & \textbf{94.8}   \\
(improve)   & {\color{teal}+5.2}\\
\bottomrule
\end{tabular}
\caption{CIFAR-100}
\label{tab:one-class-cifar100}
\end{subtable}
\hspace{2.2mm}
\begin{subtable}{0.48\linewidth}
\centering
\setlength{\tabcolsep}{8mm}
\begin{tabular}{c|ccccccccccc}
\toprule
Method   & AUROC   \\
\midrule
Rot\cite{rot}  & 65.3  \\
Rot+Trans\cite{rot} & 77.9  \\
Rot+Attn\cite{rot} & 81.6  \\
Rot+Trans+Attn\cite{rot} & 84.8  \\
Rot+Trans+Attn+Resize\cite{rot} & 85.7  \\
CSI (SOTA) \cite{csi}  & 91.6  \\
ours & \textbf{92.0}   \\
(improve)  & {\color{teal}+0.4}\\
\bottomrule
\end{tabular}
\caption{ImageNet-30}
\label{tab:one-class-imagenet30}
\end{subtable}
\caption{\textbf{One-class OOD detection.} AUROC (\%) of OOD methods on one-class (a) CIFAR-10, (b) CIFAR-100 (super-classes) and (c) ImageNet-30. The reported results on CIFAR-10 are averaged over 3 trials. Subscripts denote standard deviation, and bold ones denote the best results. The last line lists improvement of MOOD over the current SOTA.}
\label{tab:one-class}
\end{table*}

\begin{table*}[t]
\small
\centering
\setlength{\tabcolsep}{3.2mm}
\begin{subtable}{0.99\linewidth}
\centering
\begin{tabular}{c|cccccccccc}
\toprule
In-Distribution & \multicolumn{4}{c|}{CIFAR-10 $\longrightarrow $} &\multicolumn{4}{c}{CIFAR-100 $\longrightarrow $} \\
Out-of-Distribution & SVHN  & CIFAR-100  & LSUN & \multicolumn{1}{c|}{Average} & SVHN  & CIFAR-10   & LSUN & Average \\
\midrule
Baseline OOD\cite{baseline_ood} & 88.6  & 85.8 &90.7 &\multicolumn{1}{c|}{88.4} & 81.9  & 81.1 &86.6 &83.2   \\
ODIN\cite{odin}   & 96.4  & 89.6 &-   &\multicolumn{1}{c|}{93.0} & 60.9  & 77.9 &-   &69.4   \\
Mahalanobis\cite{mahalanobis}  & 99.4  & 90.5 &-   &\multicolumn{1}{c|}{95.0} & 94.5  & 55.3 &-   &74.9   \\
Residual Flows\cite{residual_flows}   & 99.1  & 89.4 &-   &\multicolumn{1}{c|}{94.3} & 97.5  & 77.1 &-   &87.3   \\
Gram Matrix\cite{gram_matrix}  & 99.5  & 79.0 &-   &\multicolumn{1}{c|}{89.3} & 96.0  & 67.9 &-   &82.0   \\
Outlier exposure\cite{outlier_exposure} & 98.4  & 93.3 &-   &\multicolumn{1}{c|}{95.9} & 86.9  & 75.7 &-   &81.3   \\
Rotation loss\cite{rot} & 98.9  & 90.9 &–   &\multicolumn{1}{c|}{94.9} & - & -   &-   &-   \\
Contrastive loss\cite{supcon} & 97.3  & 88.6 &92.8 &\multicolumn{1}{c|}{92.9} & 95.6  & 78.3 &-   &87.0   \\
CSI\cite{csi} & 97.9  & 92.2 &97.7 &\multicolumn{1}{c|}{95.9} & - & -   &-   &-   \\
SSD+ (SOTA) \cite{ssd}   & \textbf{99.9}   & 93.4 &98.4 &\multicolumn{1}{c|}{97.2} & \textbf{98.2}   & 78.3 &79.8 &85.4   \\
ours   & 99.8\tiny{$\pm$0.0} & \textbf{99.4}\tiny{$\pm$0.0} & \textbf{99.9}\tiny{$\pm$0.0} & \multicolumn{1}{c|}{\textbf{99.7}} & 96.5\tiny{$\pm$0.6} & \textbf{98.3}\tiny{$\pm$0.1} & \textbf{96.3}\tiny{$\pm$0.6} & \textbf{97.0}   \\
(improve) & {\color{gray}-0.1}   & {\color{teal}+6.0}  & {\color{teal}+1.5}  & \multicolumn{1}{c|}{{\color{teal}+2.5}} & {\color{gray}-1.7}   & {\color{teal}+20.0}   & {\color{teal}+16.5}   & {\color{teal}+11.6}\\
\bottomrule
\end{tabular}
\caption{CIFAR}
\end{subtable}

\begin{subtable}{0.99\linewidth}
\centering
\setlength{\tabcolsep}{3.8mm}
\begin{tabular}{c|ccccccccccccccc}
\toprule
In-Distribution & \multicolumn{8}{c}{ImageNet-30 $\longrightarrow $}   \\
Out-of-Distribution & Dogs  & Places365 & Flowers102 & Pets  & Food  & Caltech256 & DTD   & Average  \\
\midrule
Baseline OOD\cite{baseline_ood} & 96.7  & 90.5  & 89.7  & 95.0  & 79.8  & 90.6  & 90.1  & 90.3  \\
Contrastive loss\cite{supcon} & 95.6  & 89.7  & 92.2  & 94.2  & 81.2  & 90.2  & 92.1  & 90.7  \\
CSI (SOTA)\cite{csi}   & 98.3  & 94.0  & 96.2  & 97.4  & 87.0  & 93.2  & 97.4  & 94.8  \\
ours  & \textbf{99.4} & \textbf{98.9} & \textbf{100.0} & \textbf{99.1} & \textbf{96.6} & \textbf{99.5} & \textbf{98.9} & \textbf{98.9}  \\
(improve)   & {\color{teal}+0.9} & {\color{teal}+4.9} & {\color{teal}+3.8} & {\color{teal}+1.7} & {\color{teal}+9.6} & {\color{teal}+6.3} & {\color{teal}+1.5} & {\color{teal}+4.1}\\
\bottomrule
\end{tabular}
\caption{ImageNet-30}
\end{subtable}

\begin{subtable}{0.99\linewidth}
\centering
\setlength{\tabcolsep}{7.7mm}
\begin{tabular}{c|ccccccccccccccc}
\toprule
In-Distribution                              & \multicolumn{5}{c}{ImageNet-1k $\longrightarrow $}                                                       \\
Out-of-Distribution                             & iNaturalist        & SUN                & Places             & Textures            & Average            \\
\midrule
Baseline OOD \cite{baseline_ood}         & 87.6               & 78.3               & 76.8               & 74.5                & 79.3               \\
ODIN \cite{odin}                & 89.4               & 83.9               & 80.7               & 76.3                & 82.6               \\
Energy \cite{energy}            & 88.5               & 85.3               & 81.4               & 75.8                & 82.7               \\
Mahalanobis \cite{mahalanobis}  & 46.3               & 65.2               & 64.5               & 72.1                & 62.0               \\
GradNorm (SOTA) \cite{gradnorm} & \textbf{90.3}      & 89.0               & 84.8               & 81.1                & 86.3               \\
ours                     & 86.9               & \textbf{89.8}      & \textbf{88.5}      & \textbf{91.3}       & \textbf{89.1}      \\
(improve)                       & {\color{gray}-3.4} & {\color{teal}+0.8} & {\color{teal}+3.7} & {\color{teal}+10.2} & {\color{teal}+2.8}\\
\bottomrule
\end{tabular}
\caption{ImageNet-1k}
\end{subtable}
\caption{\textbf{Multi-class OOD detection.} AUROC (\%) of OOD detection methods on multi-class CIFAR-10, CIFAR-100, ImageNet-30 and ImageNet-1k. The reported results on CIFAR-10 and CIFAR-100 are averaged over 3 trials. Subscripts denote standard deviation, and bold ones stand for the best results. The last line lists improvement of MOOD over the current SOTA approach.}
\label{tab:multi-class}
\end{table*}


\section{Experiments}
\label{sec:exp}
In this section, we compare Masked Image Modeling for OOD detection (MOOD) with current SOTA approaches in one-class OOD detection (\cref{sec:1class}), multi-class OOD detection (\cref{sec:multi-class}), near-distribution OOD detection (\cref{sec:near-distribution}) and OOD detection with few-shot outlier exposure (\cref{sec:exposure}). Our MOOD outperforms all previous approaches on all four OOD detection tasks significantly. 

\vspace{2mm}\noindent\textbf{Experimental Configuration.} We report the commonly-used Area Under the Receiver Operating Characteristic Curve (AUROC) as a threshold-free evaluation metric for detecting OOD score. We perform experiments on (i) CIFAR-10 \cite{cifar}, which consists of 50,000 training and 10,000 testing images with 10 image classes, (ii) CIFAR-100 \cite{cifar} and CIFAR-100 (super-classes) \cite{cifar}, which consists of 50,000 training and 10,000 testing images with 100 and 20 (super-classes) image classes. respectively, (iii) ImageNet-30 \cite{imagenet}, which contains 39,000 training and 3,000 testing images with 30 image classes, and (iv) ImageNet-1k \cite{imagenet}, which contains around 120k and 50k testing images with 1k image classes.
More details of training settings are given in the Appendix.

\subsection{One-Class OOD Detection}
\label{sec:1class}
We start with the one-class OOD detection. For a given multi-class dataset of $N_c$ classes, we conduct $N_c$ one-class OOD tasks, where each task regards one of the classes as in-distribution and the remaining classes as out-of-distribution. We run our experiments on three datasets, following prior work \cite{geom, rot, goad}, of CIFAR-10, CIFAR-100 (super-classes), and ImageNet-30. 

\Cref{tab:one-class} summarizes the results, showing that MOOD outperforms current SOTA of CSI \cite{csi} on all tested cases significantly. The improvement is of 5.7\% to 94.9\% on average. The improvement is comparatively smaller on ImageNet-30 \cref{tab:one-class-imagenet30}. It is because we do not apply intermediate fine-tuning of the model on ImageNet-30. More details are shown in \cref{sec:fine-tune}. We provide the class-wise AUROC in the Appendix for detailed exhibition.



\subsection{Multi-Class OOD Detection} 
\label{sec:multi-class}
For multi-class OOD Detection, we assume that ID samples are from a specific multi-class dataset. They are tested on various external datasets as out-of-distribution. We perform MOOD on CIFAR-10, CIFAR-100, ImageNet-30 and ImageNet-1k. For CIFAR-10, We consider CIFAR-100 \cite{cifar}, SVHN \cite{svhn} and LSUN \cite{lsun} as OOD datasets. For CIFAR-100, We consider CIFAR-10 \cite{cifar}, SVHN \cite{svhn} and LSUN \cite{lsun} as OOD datasets. For ImageNet-30, OOD samples are from CUB-200 \cite{cub}, Stanford Dogs \cite{dogs}, Oxford Pets \cite{pets}, Oxford Flowers \cite{flowers}, Food-101 \cite{food}, Places-365 \cite{places}, Caltech-256 \cite{caltech}, and Describable Textures Dataset (DTD) \cite{dtd}. For ImageNet-1k, we utilize non-natural images as OOD datasets, which includes iNatualist \cite{inaturalist}, SUN \cite{sun}, places \cite{places}, Textures \cite{dtd}.

As shown in \cref{tab:multi-class}, MOOD boosts performance of current SOTA of SSD+ \cite{ssd} by 3.0\% to 97.6\% and SOTA of GradNorm \cite{gradnorm} by 2.8\% to 89.1\% on ImageNet-1k. We remark that when detecting hard (i.e., near-distribution) OOD samples on ImageNet30 and Food, MOOD still yields decent performance, while previous methods often fail. 

\begin{figure}[t]
  \centering
  \begin{subfigure}{\linewidth}
    \includegraphics[width=0.99\linewidth, trim=0 0 0 0, clip]{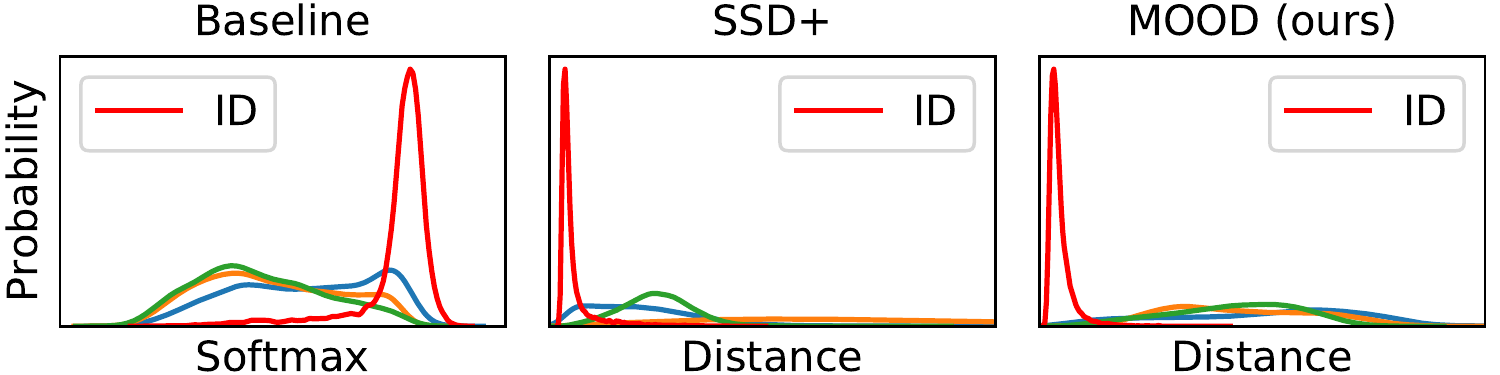}
    \caption{ID: CIFAR-10}
    \label{fig:distribution_cifar100}
  \end{subfigure}
  \\
  \begin{subfigure}{\linewidth}
    \includegraphics[width=0.99\linewidth, trim=0 0 0 0, clip]{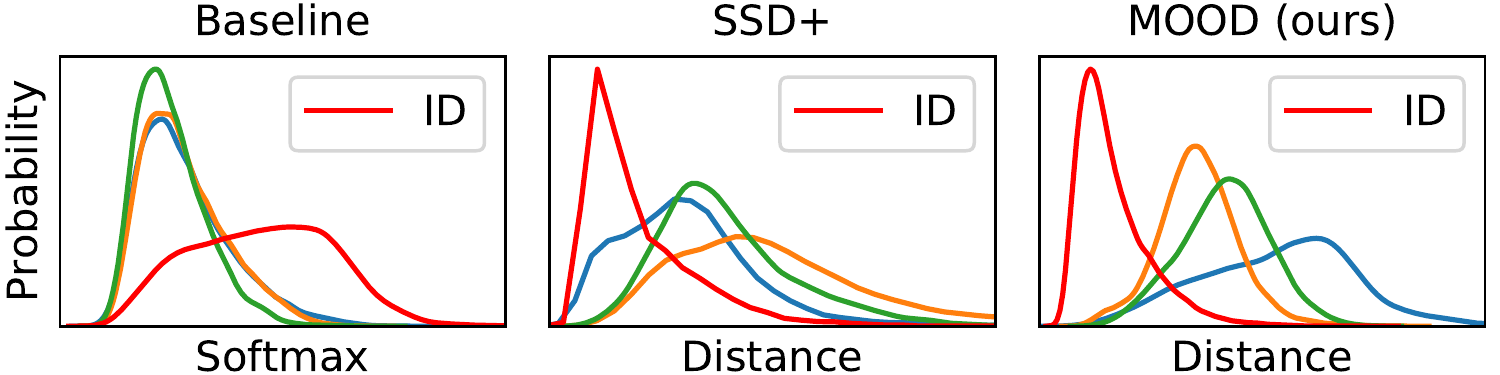}
    \caption{ID: CIFAR-100}
    \label{fig:distribution_cifar100}
  \end{subfigure}
   \\
  \begin{subfigure}{\linewidth}
    \includegraphics[width=0.99\linewidth, trim=0 0 0 0, clip]{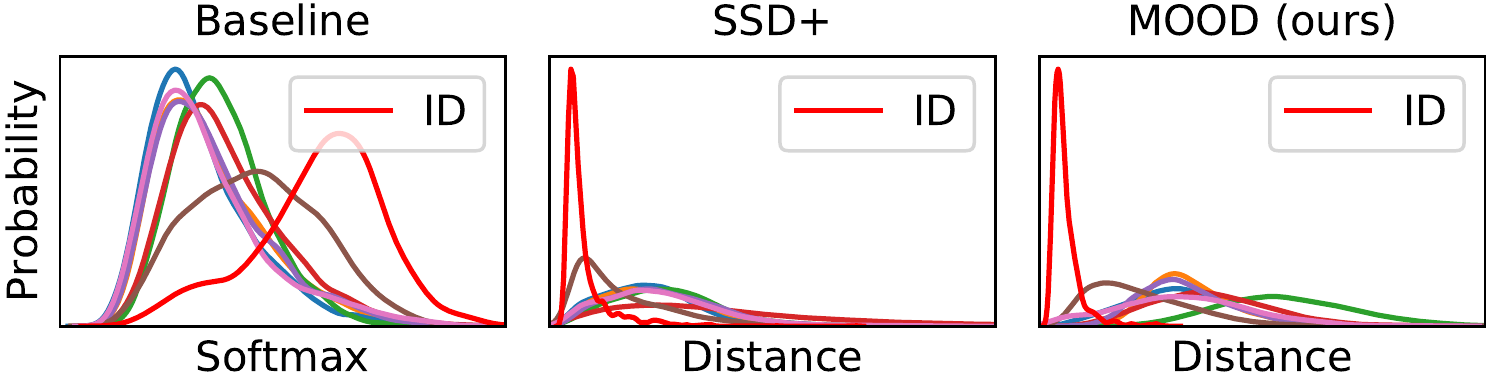}
    \caption{ID: ImageNet-30}
    \label{fig:distribution_imagenet30}
    \end{subfigure}
    \caption{Line chart to illustrate the relation between the probability distribution of test samples and OOD detection metrics on (a) CIFAR-10, (b) CIFAR-100, and (c) ImageNet-30. Each line in the sub-figures represents an OOD or ID dataset. We compare three OOD detection approaches, including baseline OOD detection, SSD+ (current SOTA, \cite{ssd} ), and our proposed MOOD. The baseline OOD detection takes the maximum softmax probabilities as its OOD detection metric, while SSD+ and MOOD both use the Mahalanobis distance as their metrics.} 
  \label{fig:distribution}
\end{figure}

\vspace{2mm}\noindent\textbf{Visualization.} In \cref{fig:distribution}, we illustrate the probability distribution of the test samples according to metrics of three OOD detection approaches: baseline OOD detection \cite{baseline_ood}, SSD+ \cite{ssd}, and MOOD. The baseline OOD detection performs softmax as its OOD detection metric, where ID samples tend to have greater value than OOD samples. MOOD and SSD+ perform the Mahalanobis distance as their metrics. 

As shown in the figure, the distance of a majority of testing ID samples to the training data is close to zero, demonstrating a similar representation of training and testing ID samples. In contrast, the distances from most OOD samples to the training data are much larger, especially on CIFAR-10 and ImageNet-30. 

Also, in \cref{fig:distribution}, we reveal that the difference in the distribution of ID and OOD samples according to MOOD is significantly larger compared with other approaches \cite{baseline_ood, ssd}. It demonstrates that MOOD can separate ID and OOD samples more clearly. In order to illustrate the appearance of images in each ID and OOD dataset, we plot several images as examples with their corresponding distances in the Appendix.


\subsection{Near-Distribution OOD Detection} 
\label{sec:near-distribution}
Compared with existing approaches on normal OOD detection tasks, SOTA results of near-distribution OOD detection is much worse -- AUROC of some ID-OOD pairs \cite{csi, ssd} is even lower than 70\%. Therefore, improving SOTA for near-OOD detection is essential for the application to work on real-world data. 

\begin{table}[t]
\small
\centering
\setlength{\tabcolsep}{2.8mm}
\begin{tabular}{cc|cccccc|cccc}
\toprule
ID & OOD  & \multicolumn{3}{c}{AUROC (\%)} \\
class & class  & CSI \cite{csi} & ours  & (improve)           \\
\midrule
Plane   & Automobile   & 74.1 & 99.0  & {\color{teal}+24.9} \\
Plane   & Ship  & 79.6 & 99.4  & {\color{teal}+19.8} \\
Plane   & Truck & 82.8 & 98.5  & {\color{teal}+15.7} \\
Bird    & Horse & 83.2 & 94.3  & {\color{teal}+11.1} \\
Cat     & Deer  & 83.3 & 92.6  & {\color{teal}+9.3}  \\
Cat     & Dog   & 67.0 & 75.5  & {\color{teal}+8.5}  \\
Cat     & Frog  & 89.6 & 92.5  & {\color{teal}+2.9}  \\
Cat     & Horse & 79.0 & 95.5  & {\color{teal}+16.5} \\
Deer    & Horse & 69.0 & 100.0 & {\color{teal}+31.0}   \\
Dog     & Deer  & 88.1 & 96.4  & {\color{teal}+8.3}  \\
Dog     & Horse & 76.6 & 95.5  & {\color{teal}+18.9} \\
Trunk   & Automobile   & 72.3 & 87.8  & {\color{teal}+15.5}\\
\midrule
\multicolumn{2}{c|}{Average} & 78.7 & 93.9 & {\color{teal}+15.2} \\
\bottomrule
\end{tabular}
\caption{\textbf{Near-distribution OOD detection} (one-class). AUROC (\%) of near-distribution pairs in one-class detection on CIFAR-10, compared with current SOTA (CSI \cite{csi}).}
\label{tab:hardood-oneclass}
\end{table}

\begin{figure}[t]
  \centering
    \includegraphics[width=0.99\linewidth, trim=0 15 0 10 clip]{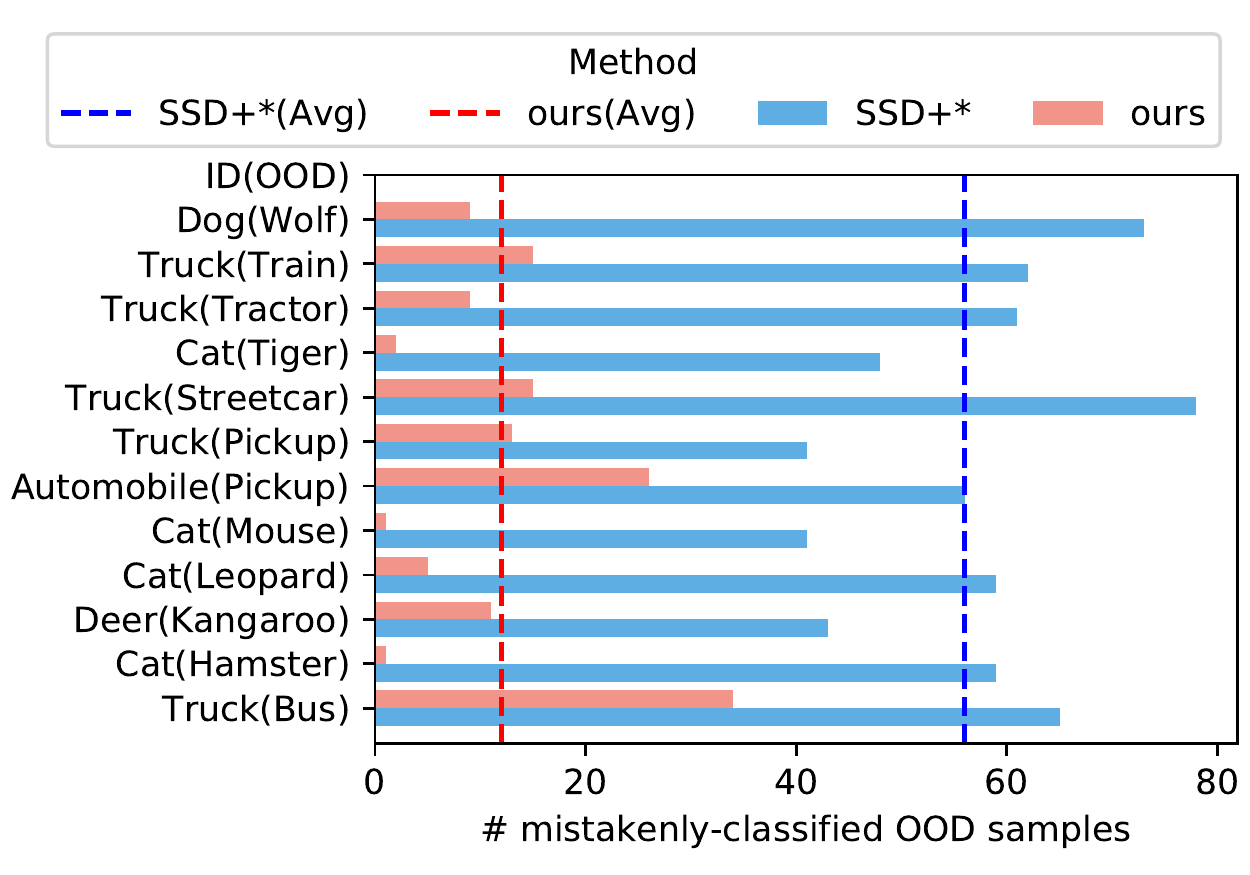}
    \caption{\textbf{Near-distribution OOD detection} (multi-class). Number of some mistakenly-classified OOD samples (when TPR = 95\%). These samples are wrongly taken as ID samples by the current SOTA of SSD+ \cite{ssd} in multi-class detection on CIFAR-10. `*' indicates SOTA.}
  \label{fig:hardood-multiclass}
\end{figure}

In \cref{tab:structure}, we have compared MOOD with the current SOTA on near-distribution CIFAR10-CIFAR100 (ID-OOD) pair, R50+ViT \cite{oodlimits}, and MOOD outperforms the latter significantly by 2.07\% to 98.30\%. In this section, we focus on the hard-detected pairs with similar semantics from \cref{sec:1class} and \cref{sec:multi-class}. 

For one-class OOD detection, we adopt 12 hard-detected ID-OOD pairs (AUROC under 90\%) from the confusion matrix of current one-class OOD detection SOTA of CSI \cite{csi}. The semantics of these ID-OOD pairs are more similar than normal ID-OOD combinations, such as trunk and car, deer and horse, etc., leading to their poor OOD detection performance. As shown in \cref{tab:hardood-oneclass}, MOOD significantly boosts the AUROC of current SOTA from 78.7\% to 93.9\%. 

For multi-class OOD detection, we examine the large mistakenly-classified value in the OOD-ID confusion matrix, which represents the number of classifying the OOD image to the category in the ID dataset. For example, when the True-Positive Rate (TPR) is 95\%, 48 testing tiger images from CIFAR-100 are classified as cat by the current multi-class OOD detection SOTA method of SSD+ \cite{ssd}, while only 2 of them are wrongly classified by MOOD. More results are shown in \cref{fig:hardood-multiclass}. For the listed 12 ID-OOD pairs, MOOD averagely reduces the number of mistakenly-classified OOD samples notably by 79\%.

\begin{table}[t]
\small
\centering
\setlength{\tabcolsep}{3.1mm}
\begin{tabular}{c|ccccccccccc}
\toprule
\multirow{2}{*}{Method}  & \# OOD samples & \multirow{2}{*}{AUROC(\%)} \\
                         & per class  &                           \\
\midrule
\multirow{5}{*}{R50+ViT (SOTA) \cite{oodlimits}} & 0          & 98.52                     \\
                         & 1          & 98.96                     \\
                         & 2          & 99.11                     \\
                         & 3          & 99.17                     \\
                         & 10         & 99.29                     \\
\midrule
ours           & 0          & \textbf{99.41}            \\
(improve)                & -          & {\color{teal}+0.12}      \\
\bottomrule
\end{tabular}
\caption{\textbf{Outlier Exposure OOD detection.} AUROC (\%) of current SOTA of R50+ViT \cite{oodlimits} for near-distribution OOD detection and MOOD. SOTA utilizes up to 10 known OOD samples per class for detection, while ours do not include any OOD samples.}
\label{tab:exposure}
\end{table}

\subsection{OOD Detection with Outlier Exposure} 
\label{sec:exposure}
Several representative OOD detection methods \cite{ssd, oodlimits} utilize OOD samples to improve the performance in extra stages. We note they are not included in our work because we generally believe that exposure of OOD samples violates the original intention of OOD detection. 

In \cref{tab:exposure}, we compare MOOD with current SOTA \cite{oodlimits} for near-distribution OOD detection with up to 10 OOD samples per class. We surprisingly find that MOOD works better in terms of AUROC than current SOTA \cite{oodlimits}, even though we do not include any OOD samples for detection. The outstanding performance of MOOD demonstrates that an effective pretext task is already sufficient for producing a distinguishable representation that OOD detection requires. Thus, there is no need to include extra OOD samples.


\section{Conclusion}
In this paper, we have extensively explored the effect of multiple contributors for OOD detection and observed that reconstruction-based pretext tasks have the potential to provide effective priors for OOD detection to learn the real data distribution of the ID dataset. Specifically, we take the Masked Image Modeling pretext task for our OOD detection framework (MOOD). We perform MOOD on one-class OOD detection, multi-class OOD detection, near-distribution OOD detection, and few-shot outlier exposure OOD detection -- MOOD all achieve new SOTA results, although we do not include any OOD samples for detection.

\section{Acknowledgement}
This work is partially supported by Shenzhen Science and Technology Program KQTD20210811090149095.


{
\bibliographystyle{ieee_fullname}
\bibliography{egbib}
}

\end{document}